\documentclass[letterpaper, 10 pt, conference]{ieeeconf}  
\IEEEoverridecommandlockouts                             
\usepackage{epsfig} 
\usepackage{mathptmx} 
\usepackage{times} 
\usepackage{amsmath} 
\usepackage{amssymb}  
\usepackage[table]{xcolor}
\usepackage{booktabs}
\usepackage{makecell}
\usepackage{multirow}
\usepackage[T1]{fontenc}

\usepackage{multirow}
\usepackage{float}
\usepackage{bm}
\usepackage{hyperref}
\definecolor{custompink}{RGB}{239,118,186} 
\hypersetup{
    colorlinks=true, 
    linkcolor=blue,  
    citecolor=green, 
    filecolor=magenta, 
    urlcolor=custompink     
}

\title{\LARGE \bf
Coastal Underwater Evidence Search System with \\
Surface-Underwater Collaboration}

\author{Hin Wang Lin$^{\dag, 1}$, Pengyu Wang$^{\dag, 1}$$^*$, Zhaohua Yang$^{1}$, Ka Chun Leung$^{2}$, \\ Fangming Bao$^{1}$, Ka Yu Kui$^{1}$, Jian Xiang Erik Xu$^{3}$ and Ling Shi$^{1}$, \textit{Fellow, IEEE}%
\thanks{$^{\dag}$ Co-first authors. $^*$ Corresponding author.}
\thanks{$^{1}$Hin Wang Lin, Pengyu Wang, Zhaohua Yang, Ka Chun Leung, Fangming Bao, Ka Yu Kui and Ling Shi are with the Department of Electronic and Computer Engineering, Hong Kong University of Science and Technology, Hong Kong SAR. {\tt\small \{hwlinaa, pwangat, zyangcr, fbao, kykui\}@connect.ust.hk, eesling@ust.hk}}%
\thanks{$^{2}$Ka Chun Leung is with the Department of Computer Science and Engineering, Hong Kong University of Science and Technology, Hong Kong SAR. {\tt\small kcleungax@connect.ust.hk}}%
\thanks{$^{3}$Jian Xiang Erik Xu is with the King George V School, Hong Kong SAR. {\tt\small xujianxiang338@gmail.com}}%
}

\begin{document}

\maketitle
\thispagestyle{empty}
\pagestyle{empty}

\begin{abstract}
The Coastal underwater evidence search system with surface-underwater collaboration is designed to revolutionize the search for artificial objects in coastal underwater environments, overcoming limitations associated with traditional methods such as divers and tethered remotely operated vehicles. Our innovative multi-robot collaborative system consists of three parts: an autonomous surface vehicle as a mission control center, a towed underwater vehicle for wide-area search, and a biomimetic underwater robot inspired by marine organisms for detailed inspections of identified areas. We conduct extensive simulations and real-world experiments in pond environments and coastal fields to demonstrate the system’s potential to surpass the limitations of conventional underwater search methods, offering a robust and efficient solution for law enforcement and recovery operations in marine settings. (Video\footnote{\href{https://youtu.be/tEtTLpGrO2c}{https://youtu.be/tEtTLpGrO2c}})

\end{abstract}

\section{Introduction} Underwater evidence such as weapons, clothing, electronic devices, and human remains is crucial for forensic analysis, offering more reliable information than items exposed on land~\cite{qiu2019underwater}. These undisturbed artifacts provide essential insights into criminal activities, maritime accidents, and other underwater investigations~\cite{plueddemann2011autonomous}. However, locating and recovering underwater evidence is fraught with challenges, including poor visibility, complex terrains, and significant risks to human divers~\cite{rutledge2018intelligent}. Traditional methods, including divers and ROVs, face limitations such as human risk, poor underwater visibility, and difficulties navigating complex terrains. This paper presents the Coastal Underwater Evidence Search System (CUES-S), which integrates autonomous surface vehicles (ASVs) and biomimetic underwater robots to enhance the efficiency and safety of underwater evidence search and recovery operations.

Underwater robotics has seen significant advancements driven by the need to explore and perform tasks in complex marine environments. Remotely Operated Vehicles (ROVs) and Autonomous Underwater Vehicles (AUVs) are the primary platforms used in underwater operations. ROVs, such as BlueROV2 and Seaeye Falcon, are renowned for their agility and high-definition visual capabilities. However, their tethered nature and propeller-driven movement often lead to entanglement and sediment disturbance, limiting their efficiency in complex terrains \cite{park2015multi}.  

\begin{figure}[t!] 
\centerline{\includegraphics[width=1\linewidth]{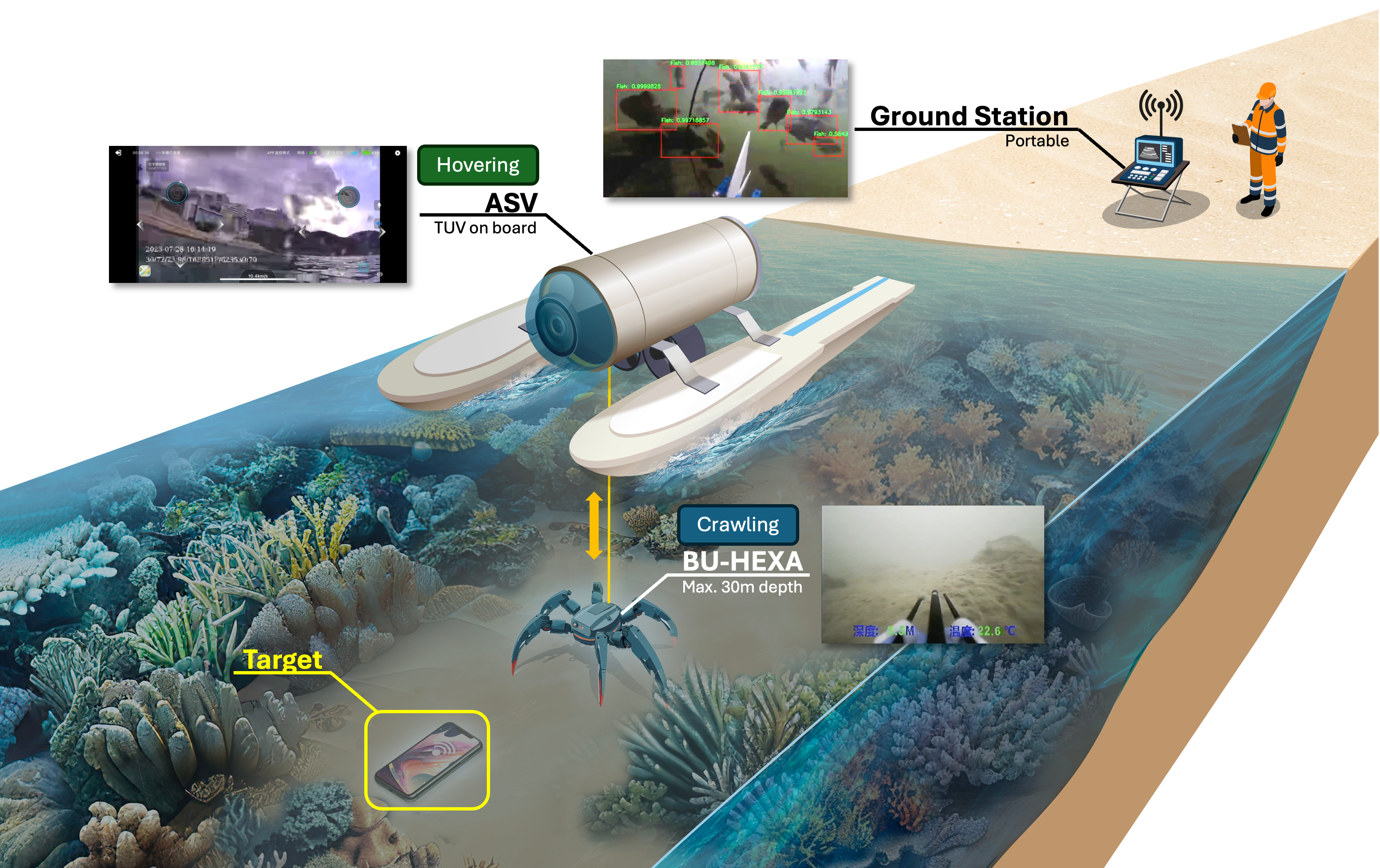}} 
\caption{Schematic diagram of CUES-S operation.} 
\label{fig:stages} 
\end{figure}

AUVs, like those developed by Qingdao Pengpai, offer greater autonomy and are ideal for large-scale surveys. However, they face challenges in precise object retrieval and maneuvering in confined spaces \cite{pengpai_auv}. Seabed crawling robots, which employ robust tracked movement systems, excel in navigating pliable substrates but struggle in rocky or coral-rich environments. The need for versatile, stable, and minimally invasive underwater robots has prompted exploration into biomimetic designs. Inspired by marine organisms such as fishes, soft robots offer superior stability and adaptability in navigating complex underwater terrains \cite{qu2024recent}.  

The integration of multiple robotic platforms into a cohesive system for underwater search and recovery has shown promise in enhancing the efficiency and flexibility of search operations. Multi-agent system (MAS) approaches have demonstrated the benefits of cooperative AUV navigation using a single maneuvering surface craft \cite{fallon2010cooperative}, and algorithms for task allocation and coordination among multiple USVs and AUVs have been developed \cite{wang2023cooperative}. In~\cite{wang2022quadrotor}, the authors propose
an efficient collaboration planning framework for a heterogeneous robot system.
This framework enhanced scalability and robustness but demanded significant
computational resources and complex algorithmic development. Lindsay \textit{et al.}~\cite{lindsay2022collaboration} proposed a collaborative system integrating three robots to achieve multi-domain sensing
and situational awareness on partially submerged targets. Despite advancements, robust communication and coordination protocols remain a significant challenge.  

Accurate detection and identification of underwater objects rely heavily on advanced imaging and sensing technologies. Li \textit{et al.}~\cite{li2021development} developed a buoy-borne underwater imaging system for in situ
mesoplankton monitoring of coastal waters, which offers insights for underwater evidence search systems. High-resolution sonar and video imaging systems, enhanced with machine learning algorithms, have been developed to analyze data and identify evidence accurately \cite{koutalakis2022river}. However, turbid waters and complex terrains present ongoing challenges. Multi-sensor fusion approaches, combining sonar, video, and magnetic sensors, offer a comprehensive view and improved detection capabilities but necessitate sophisticated data processing algorithms \cite{pengpai_auv}.

Our primary contributions include the development of a multi-robot collaborative system that integrates an autonomous surface vehicle (ASV), a towed underwater vehicle (TUV), and a biomimetic underwater hexapod robot (BU-HEXA). This innovative system addresses key challenges in underwater evidence search by enhancing maneuverability, stability, and detection capabilities. The ASV serves as a mission control center, ensuring precise navigation and robust communication, while the TUV covers wide-area searches efficiently. The BU-HEXA, inspired by marine organisms, excels in detailed inspections of identified areas.

The remainder of this paper is organized as follows: Section II details the mission architecture of the CUES-S, outlining the stages from pre-mission planning to mission conclusion. Section III delves into the system design, describing the components and functionalities of the ASV, TUV, and BU-HEXA. Section IV presents the experimental results, evaluating the performance of the CUES-S in various controlled and real-world environments. Finally, Section V concludes the paper, summarizing our findings and discussing future directions for enhancing the system’s capabilities and extending its applications.

\section{Mission Architecture} 
The mission architecture of the CUES-S ensures seamless integration and efficient task execution through the coordinated operation of the ASV, TUV, and the biomimetic underwater hexapod robot (BU-HEXA). The mission planning and execution for underwater search and recovery operations encompass several critical stages, as shown in Fig. \ref{fig:stages}, starting with pre-mission planning.

\subsection{Pre-Mission Planning and Wide-Area Search} 
This phase involves defining mission objectives, such as identifying specific objects and collecting environmental data, selecting the target area, and developing autonomous search patterns and routes for the ASV and underwater vehicles. Deployment begins with launching the ASV to the designated area via a base station or mother ship, followed by deploying the TUV using the ASV’s winch system.

During the wide-area search phase, the TUV conducts an initial survey, capturing high-resolution imagery and environmental data. Real-time analysis is performed using the YOLO algorithm to identify potential areas of interest. The integration of machine learning techniques into the data analysis pipeline significantly enhances the identification accuracy and reduces the time required for manual inspection.

\subsection{Detailed Inspections} 
The next stage involves detailed inspections, where the BU-HEXA is deployed to closely investigate identified areas, navigating complex terrains and capturing detailed imagery with its advanced sensor suite. The BU-HEXA’s sensors and tools are used to confirm the presence of objects of interest and carry out necessary recovery tasks, such as attaching markers or retrieving objects. The hexapod's biomimetic design ensures it can maneuver effectively in confined and complex environments, providing high-resolution visual and sensory data.

\subsection{Environmental Data Collection} 
Throughout the mission, environmental sensors collect data on water quality, temperature, and other parameters, which are transmitted back to the ASV and base station for analysis and record-keeping. This data is crucial for understanding the environmental conditions and potential impacts on the evidence recovery process. The multi-sensor approach allows for comprehensive environmental monitoring and assessment, aiding in the development of strategies for minimizing ecological disturbances.

\subsection{Mission Conclusion} 
The mission concludes with the retrieval of the BU-HEXA and TUV using the ASV’s winch system, followed by the return of the ASV to the base station or mother ship. Post-mission analysis of the collected data, imagery, and any recovered objects is conducted to achieve mission objectives and provide insights for future missions. This comprehensive approach ensures that CUES-S effectively meets the technical requirements of underwater search and recovery while contributing valuable environmental data for research and conservation efforts.

\section{System Design} 
The entire system consists of three parts: ASV, TUV, and BU-HEXA, and their relationship is shown in Fig.~\ref{fig:Wiring diagram of ASV, TUV, BU-HEXA, and ground station}.

\begin{figure}[htbp] 
\centerline{\includegraphics[width=1\linewidth]{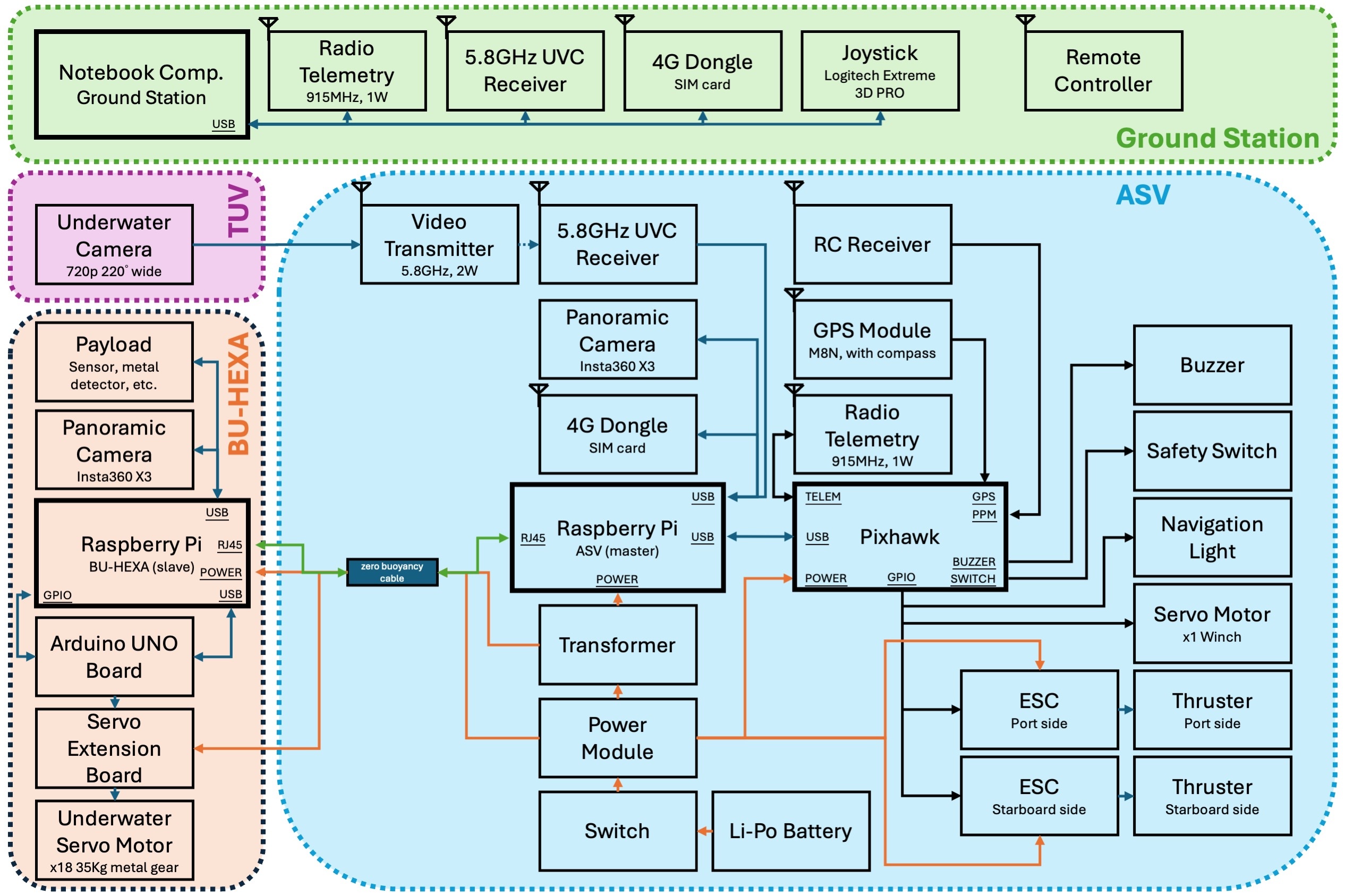}} 
\caption{Wiring diagram of ASV, TUV, BU-HEXA, and ground station.} 
\label{fig:Wiring diagram of ASV, TUV, BU-HEXA, and ground station} 
\end{figure}
\subsection{Autonomous Surface Vehicle} 
\begin{figure}[htbp] 
\centerline{\includegraphics[width=1\linewidth]{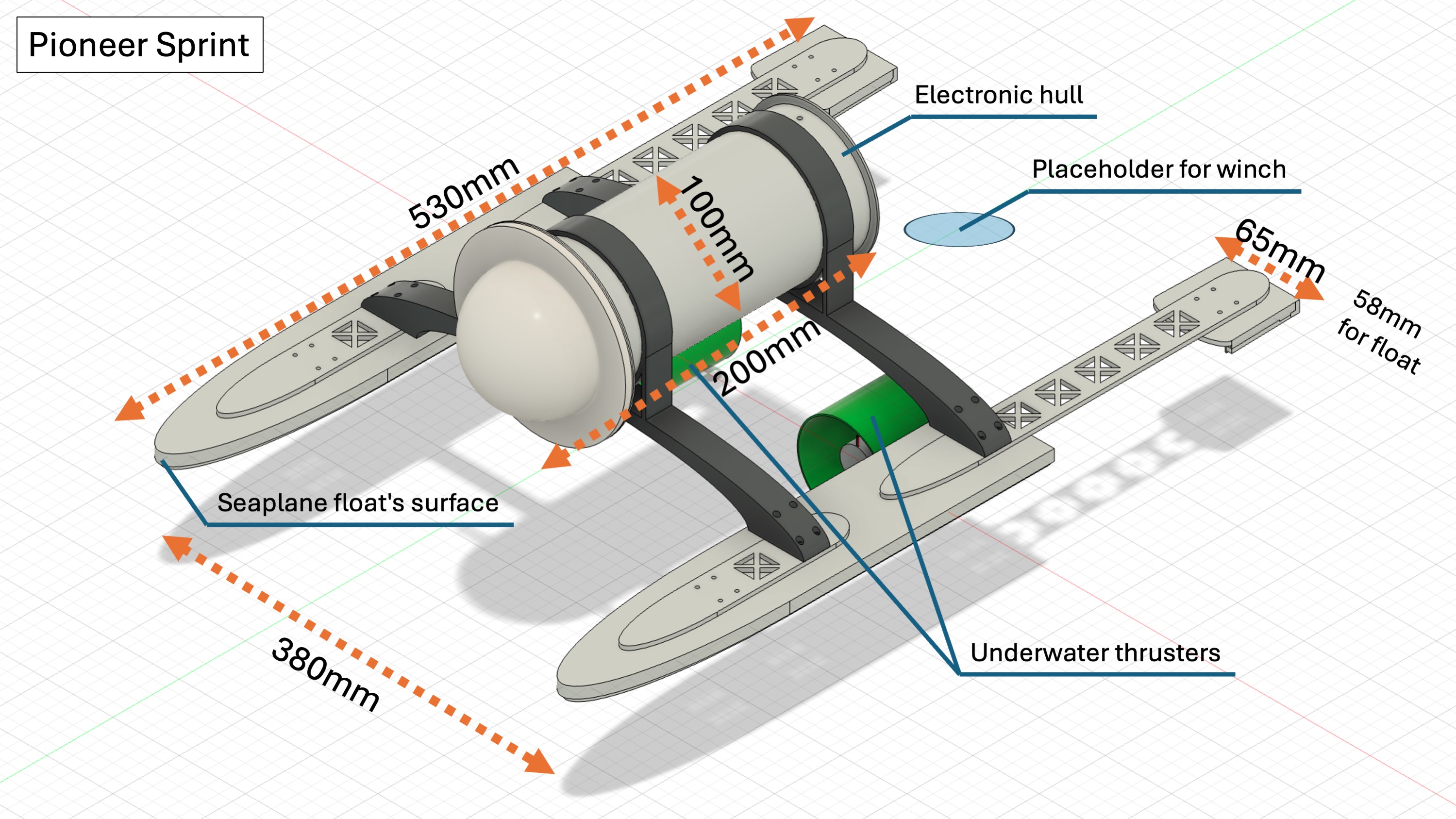}} 
\caption{3D design sketch of the Pioneer Sprint.} 
\label{fig:3D design sketch of the Pioneer Sprint} 
\end{figure}

The ASV serves as the mission control center, managing the deployment, operation, and retrieval of underwater robots, as shown in Fig.~\ref{fig:3D design sketch of the Pioneer Sprint}. It is equipped with GPS for precise positioning, autonomous navigation, and a robust communication link with the base station. The ASV's control systems include a flight controller and an onboard minicomputer for real-time data processing and mission management.  

The ASV's propulsion system consists of two brushless thrusters, providing differential thrust for precise maneuverability. The hull design is inspired by seaplane floats, offering stability and reduced drag. The system is modular, allowing for twin-hulled and four-hulled configurations to accommodate different loads. The electronic components are housed within a waterproof enclosure, ensuring reliable operation in harsh marine environments. The ASV's winch system includes a 30-meter floating cable, enabling the deployment and retrieval of underwater vehicles. 

To analyze the performance of ASV, we first define two frames: body-fixed coordinate system \(\mathbf{b} = (x_\mathbf{b}, y_\mathbf{b}, z_\mathbf{b})\) attached to the ASV and navigation coordinate system \(\mathbf{n} = (x_\mathbf{n}, y_\mathbf{n}, z_\mathbf{n})\) fixed to the earth. The location of the ASV in the navigation frame is denoted by \(x_s, y_s, z_s\), and the orientations are described by \(\phi_s, \theta_s, \psi_s\). In the body-fixed frame, the speeds along the \(x_\mathbf{b}-, y_\mathbf{b}-, z_\mathbf{b}-\) axes are denoted by \(u_s, v_s, w_s\), while the angular velocities around these axes are denoted by \(p_s, q_s, r_s\). The kinematics equation for ASV is written as: 

\begin{equation}
\begin{bmatrix}
    \dot{x}_s \\
    \dot{y}_s \\
    \dot{\psi}_s
\end{bmatrix}
=
\begin{bmatrix}
    \cos \psi_s & -\sin \psi_s & 0 \\
    \sin \psi_s &  \cos \psi_s & 0 \\
    0           &  0           & 1
\end{bmatrix}
\begin{bmatrix}
    u_s \\
    v_s \\
    r_s
\end{bmatrix}.
\end{equation}
The dynamics model for ASV is described as:
\begin{equation}
\begin{aligned}
&\begin{bmatrix}
    m_{11} & 0 & 0 \\
    0 & m_{22} & 0 \\
    0 & 0 & m_{33}
\end{bmatrix}
\begin{bmatrix}
    \dot{u}_s \\
    \dot{v}_s \\
    \dot{r}_s
\end{bmatrix} \\
&\quad+
\begin{bmatrix}
    0 & -m_{33}r_s & m_{22}v_s \\
    m_{33}r_s & 0 & -m_{11}u_s \\
    -m_{22}v_s & m_{11}u_s & 0
\end{bmatrix}
\begin{bmatrix}
    u_s \\
    v_s \\
    r_s
\end{bmatrix}
=
\begin{bmatrix}
    X \\
    Y \\
    N
\end{bmatrix},
\end{aligned}
\end{equation}
where \(m_{11}\) represents the total mass in the \(x_b\) direction, including the added mass, \(m_{22}\) represents the total mass in the \(y_b\) direction, including the added mass, and \(m_{33}\) represents the total moment of inertia about the yaw axis, including the added moment of inertia, \(X, Y, N\) are the force and moment of \(x_b-, y_b-\) axes and yaw.

The ASV is equipped with a navigation control system that includes a 4G communication module and Pixhawk flight control, which features GPS and video transmission capabilities. This setup enhances the ASV’s ability to operate remotely with high precision. Additionally, the ASV can be operated using a custom interface that allows users to start and stop the automatic hover function, ensuring that the ASV maintains its position accurately during operations.

\subsection{Biomimetic Underwater Hexapod} 
\begin{figure}[htbp] 
\centerline{\includegraphics[width=1\linewidth]{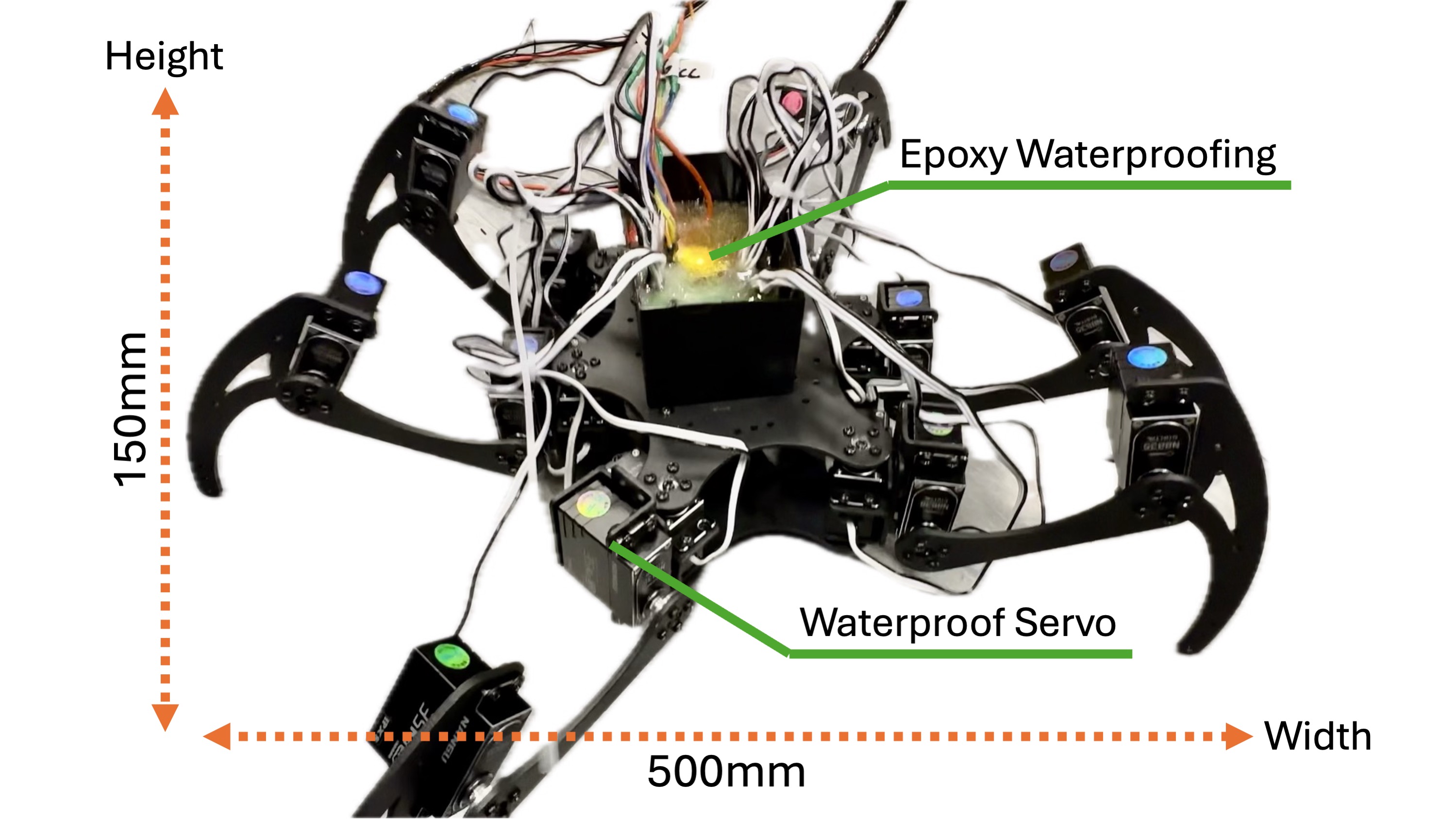}} 
\caption{Bionic Underwater Hexapod.} 
\label{fig:Bionic Underwater Hexapod} 
\end{figure}

Inspired by marine organisms like crabs, the BU-HEXA is designed for detailed inspections of identified areas. It features a six-legged design for agile and stable movement and is equipped with an integrated sensor suite, including a 360-degree camera, metal detectors, and acoustic positioning systems, as shown in Fig. \ref{fig:Bionic Underwater Hexapod}.

\begin{figure}[htbp] 
\centerline{\includegraphics[width=1\linewidth]{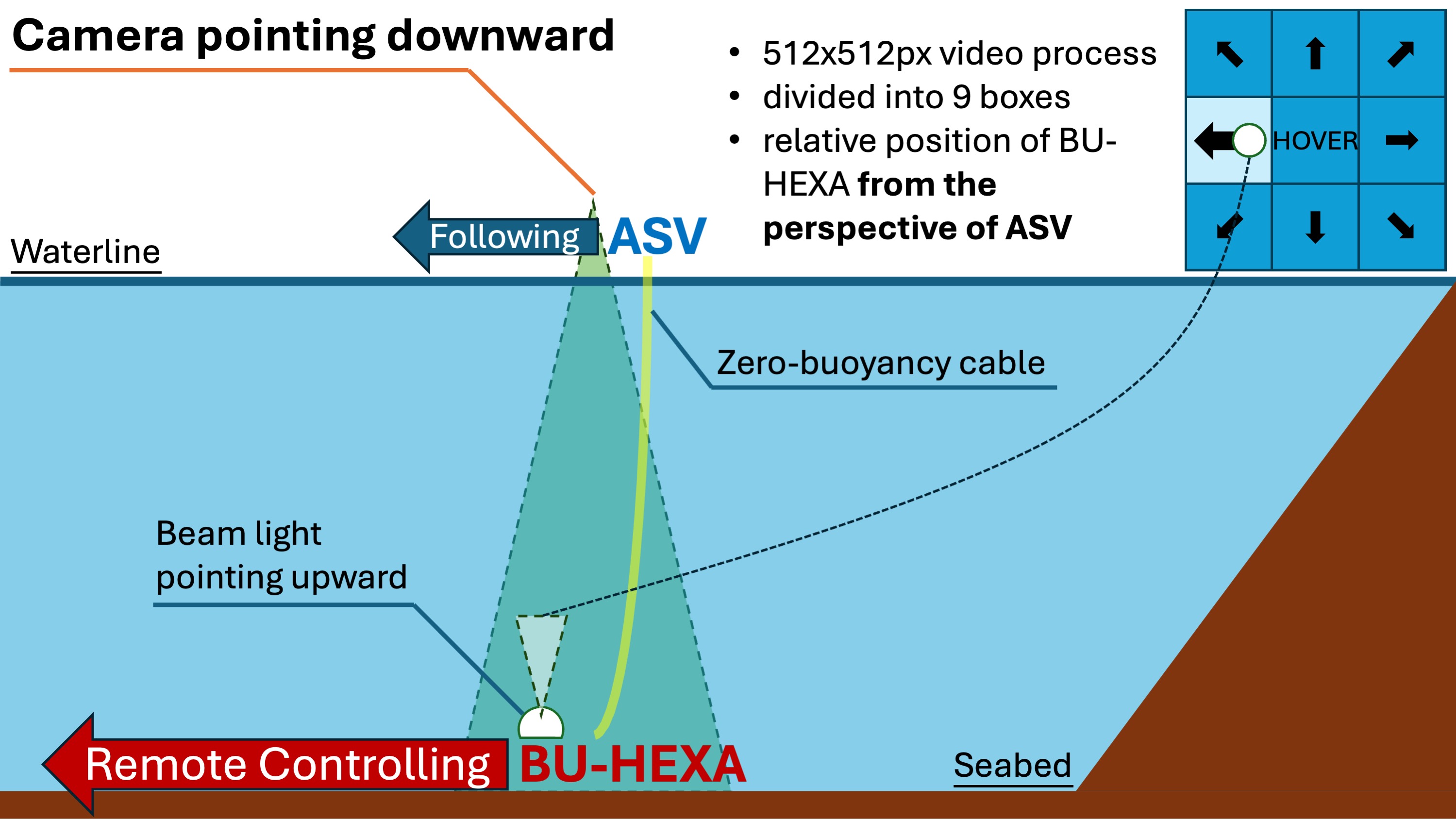}} 
\caption{Collaborative operation framework of ASV and BU-HEXA.} 
\label{fig:Collaborative operation framework of ASV and BU-HEXA} 
\end{figure}

The hexapod maintains a tethered connection to the ASV, allowing for continuous power supply and seamless data transfer. Besides, the collaboration process between the ASV and BU-HEXA is shown in Fig.~\ref{fig:Collaborative operation framework of ASV and BU-HEXA}.

The hexapod's locomotion is modeled after crab movement, providing superior stability and maneuverability on uneven seabeds. Each leg is independently controlled by waterproof servo motors, enabling precise movements. The control system utilizes advanced algorithms for trajectory planning and stability analysis, ensuring effective navigation in complex underwater terrains. The BU-HEXA's design also includes features for minimizing environmental impact, such as low-noise operation and non-invasive sensors. Specifically, in the stance phase, the locomotion is described as:
\begin{equation}
p_{j,\text{stance}}(t) = p_{j,0} + v_\text{stance} t,
\end{equation}
and in the swing phase the locomotion is described as:
\begin{equation}
p_{j,\text{swing}}(t) = p_{j,\text{stance}}(t_\text{end}) + v_\text{swing} (t - t_\text{end}).
\end{equation}
Each leg of BU-HEXA can be modeled using Denavit-Hartenberg (DH) parameters, which describe the relationship between joint angles and the position of the end effector (foot). For a single leg, the position of the end effector in the Cartesian coordinate system can be described by the forward kinematics equations:
\begin{equation}
p_{\text{foot}} = T_1(\theta_1) \cdot T_2(\theta_2) \cdot T_3(\theta_3) \cdot p_{\text{end}},
\end{equation}
where $T_i(\theta_i)$ are the transformation matrices for each joint $i$ with joint angle $\theta_i$, and $p_{\text{end}}$ is the position vector of the end effector in the local coordinate system of the last joint. Inverse kinematics involves solving for the joint angles $\theta_i$ given the desired position $p_{\text{foot}}$ of the end effector:
\begin{equation}
\theta_1 = \text{atan2}(y_{\text{foot}}, x_{\text{foot}}),
\end{equation}
\begin{equation}
\theta_2 = \text{acos}\left(\frac{d^2 + z_{\text{foot}}^2 - l_1^2 - l_2^2}{2 l_1 l_2}\right),
\end{equation}
\begin{equation}
\theta_3 = \text{atan2}\left(z_{\text{foot}}, d\right) - \text{atan2}\left(l_2 \sin(\theta_2), l_1 + l_2 \cos(\theta_2)\right),
\end{equation}
where $d = \sqrt{x_{\text{foot}}^2 + y_{\text{foot}}^2}$, and $l_1$ and $l_2$ are the lengths of the first and second leg segments, respectively. 

Waterproofing is a critical aspect of the BU-HEXA design. Each electronic component, including the servo motors and control boards, is encased in a waterproof sealant to prevent water ingress. The hexapod is also equipped with a small floating ball at the connector part to prevent the cable from tangling with the legs, ensuring smooth operation. The BU-HEXA's sensor suite includes a 720p underwater camera with a 195-degree wide-angle lens and infrared LEDs for enhanced visibility in dark or murky environments.

\subsection{Towed Underwater Vehicle} The TUV conducts wide-area searches and captures high-resolution imagery and environmental data. Its hydrodynamic design ensures stability and efficient navigation. The TUV operates tethered to the ASV, providing a continuous data flow and real-time video capture. The dynamics of the towed body can be described by Newton's second law, considering the hydrodynamic forces, added mass, and the tension in the towline:
\begin{equation}
F_b + T = (m_b + m_{b,a}) \frac{dv_b}{dt},
\end{equation}
where $T$ is the tension force vector in the towline acting on the towed body, $F_b$ includes hydrodynamic forces, gravitational forces, and other external forces acting on the towed body, $m_b$ is the mass of the towed body, $m_{b,a}$ is the added mass, representing the inertia of the water displaced by the towed body. Moreover, the lift and drag forces acting on the hydrofoil-equipped towed body can be determined using the following equations:
\begin{equation}
L = \frac{1}{2} \rho V^2 S C_L,
\end{equation}
where $L$ is the lift force, $\rho$ represents water density, $V$ is flow velocity relative to the hydrofoil, $S$ is the planform area of the hydrofoil, $C_L$ is lift coefficient, which depends on the shape of the hydrofoil and the angle of attack. The drag force is given by:
\begin{equation}
D = \frac{1}{2} \rho V^2 S C_D,
\end{equation}
where $D$ is drag force, $C_D$ is drag coefficient, which depends on the hydrofoil shape and surface roughness.

\begin{figure}[htbp] 
\centerline{\includegraphics[width=1\linewidth]{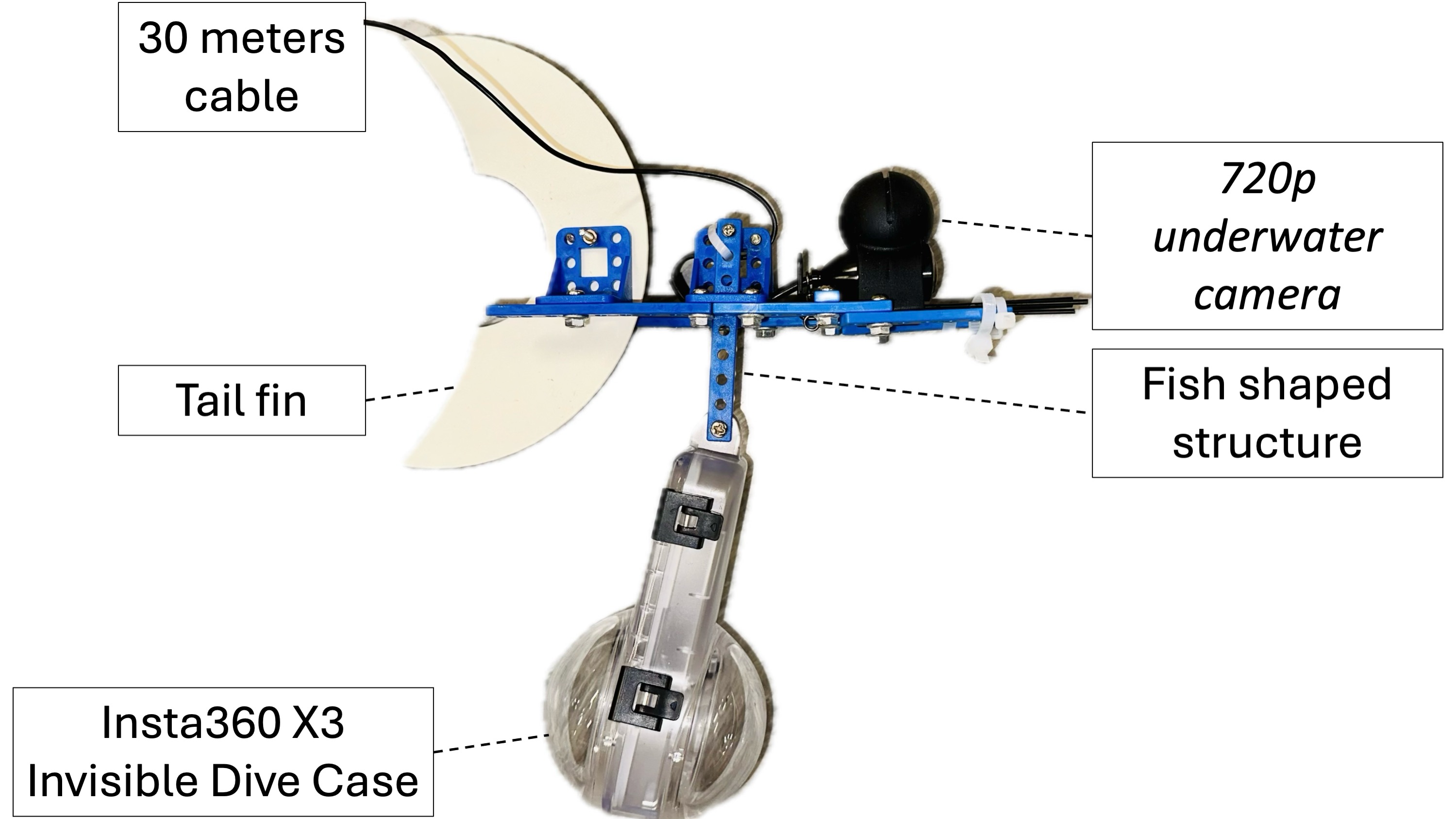}} 
\caption{Towed Underwater Vehicle.} 
\label{fig:TUV} 
\end{figure}

The TUV is equipped with a 720p underwater camera, providing high-quality video feeds for real-time image analysis, as shown in Fig.~\ref{fig:TUV}. The camera is designed to operate under high pressure, ensuring reliable performance in deep water conditions. The TUV's streamlined shape and control surfaces are optimized for minimal drag and maximum stability. The integration of advanced imaging technologies and machine learning algorithms enhances the TUV's ability to detect and classify underwater objects accurately. The TUV is designed with a robust tethering system to ensure reliable power and data transmission. This tether provides both mechanical support and a communication pathway, crucial for the TUV’s operation in challenging underwater environments. The camera system is designed to withstand high-pressure environments, ensuring consistent performance in deeper waters.

\subsection{Control Architecture}

The control architecture of the Coastal Underwater Evidence Search System (CUES-S) integrates advanced algorithms to ensure precise navigation and effective coordination among the autonomous surface vehicle (ASV), towed underwater vehicle (TUV), and biomimetic underwater hexapod robot (BU-HEXA). This section provides an overview of the control methodologies applied, specifically focusing on PID control and the Extended Kalman Filter (EKF).

\subsubsection{PID Control}

Proportional-Integral-Derivative (PID) control is employed to manage the precise movements of the ASV and BU-HEXA. This control strategy combines three components:

The PID control formula is given by:
\[ u(t) = K_p e(t) + K_i \int e(t) \, dt + K_d \frac{de(t)}{dt} \]
where \( u(t) \) is the control input, \( e(t) \) is the error, and \( K_p \), \( K_i \), \( K_d \) are the proportional, integral, and derivative gains, respectively.

\subsubsection{Extended Kalman Filter (EKF)}

The EKF is utilized for state estimation and sensor fusion, particularly for the ASV's navigation system. The EKF helps in estimating the vehicle's position, velocity, and orientation by integrating data from various sensors, such as GPS, IMUs, and compasses.

The EKF operates in two main steps:

\begin{itemize}
    \item \textbf{Prediction Step:} This step uses the vehicle's dynamic model to predict the next state based on the current state and control inputs.
    \[ \hat{x}_{k|k-1} = f(\hat{x}_{k-1}, u_{k-1}) \]
    \[ P_{k|k-1} = F_{k-1} P_{k-1} F_{k-1}^T + Q_k \]
    where \( \hat{x}_{k|k-1} \) is the predicted state, \( P_{k|k-1} \) is the predicted covariance, \( F_{k-1} \) is the Jacobian matrix of the dynamic model, and \( Q_k \) is the process noise covariance.
    \item \textbf{Update Step:} This step updates the predicted state based on the new measurements.
    \[ K_k = P_{k|k-1} H_k^T (H_k P_{k|k-1} H_k^T + R_k)^{-1} \]
    \[ \hat{x}_k = \hat{x}_{k|k-1} + K_k (z_k - h(\hat{x}_{k|k-1})) \]
    \[ P_k = (I - K_k H_k) P_{k|k-1} \]
    where \( K_k \) is the Kalman gain, \( z_k \) is the measurement, \( H_k \) is the Jacobian matrix of the measurement model, and \( R_k \) is the measurement noise covariance.
\end{itemize}

This control architecture ensures that the CUES-S can navigate complex underwater terrains with high precision and reliability, providing robust performance for evidence search and recovery operations.

\section{Experimental Results} 
\subsection{ASV Stability and Navigation Test} 
The first experiment was designed to evaluate the stability and navigation capabilities of the ASV under different sea conditions. The ASV, equipped with a 4G communication module and Pixhawk flight control system, was tested in controlled environments, including a lake and coastal sea areas.

\begin{figure}[htbp] \centerline{\includegraphics[width=1\linewidth]{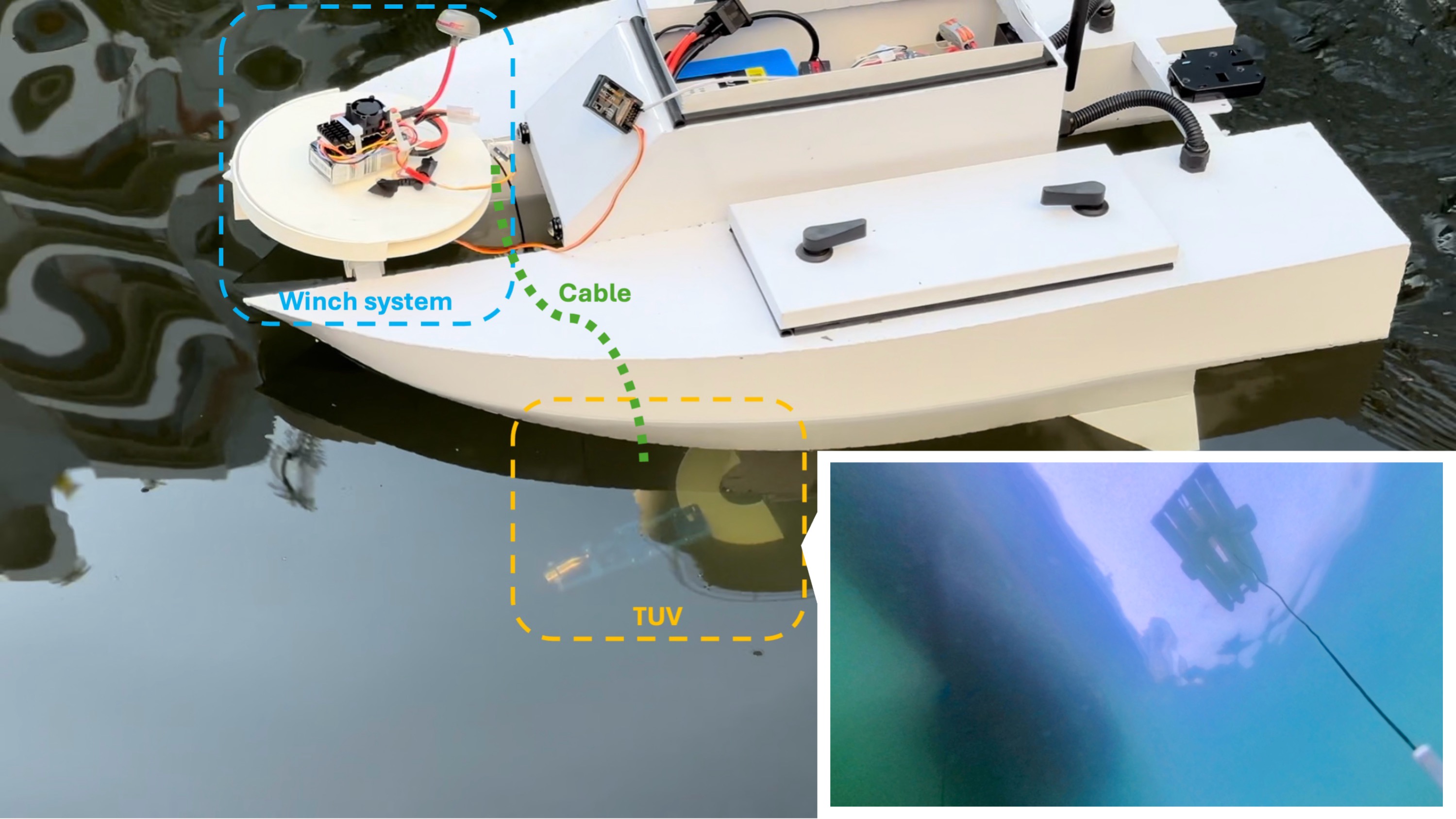}} 
\caption{The Pioneer Sprint ASV conducts trials on the lake.} 
\label{fig:lake_trial} 
\end{figure}

The field trial was conducted over two days. On the first day, the ASV was deployed in a calm lake environment to establish baseline performance metrics, as shown in Fig. \ref{fig:lake_trial}. The ASV followed a pre-defined path, maintaining a constant speed and altitude. The navigation control system’s real-time video feedback and telemetry data were recorded and analyzed to assess the ASV’s stability and maneuverability.

On the second day, the ASV was tested in coastal sea conditions to evaluate its performance in more challenging environments with stronger currents and waves. The ASV followed a similar pre-defined path as in the lake trial, and the same parameters were measured. The results from the lake trial indicated that the ASV maintained stable navigation and precise maneuverability, with minimal deviations from the planned path. The coastal sea trial results demonstrated the ASV’s ability to handle more dynamic conditions, maintaining stability and effective control despite the presence of waves and currents.

To further evaluate the ASV’s capabilities, an additional test was conducted to assess the loiter mode, which allows the ASV to hover in a specific location. This test aimed to validate the ASV’s ability to maintain a fixed position in the presence of environmental disturbances such as currents and waves. The ASV was deployed in a coastal sea area and instructed to loiter at a designated point.

\begin{figure}[htbp] 
\centerline
{\includegraphics[width=1\linewidth]{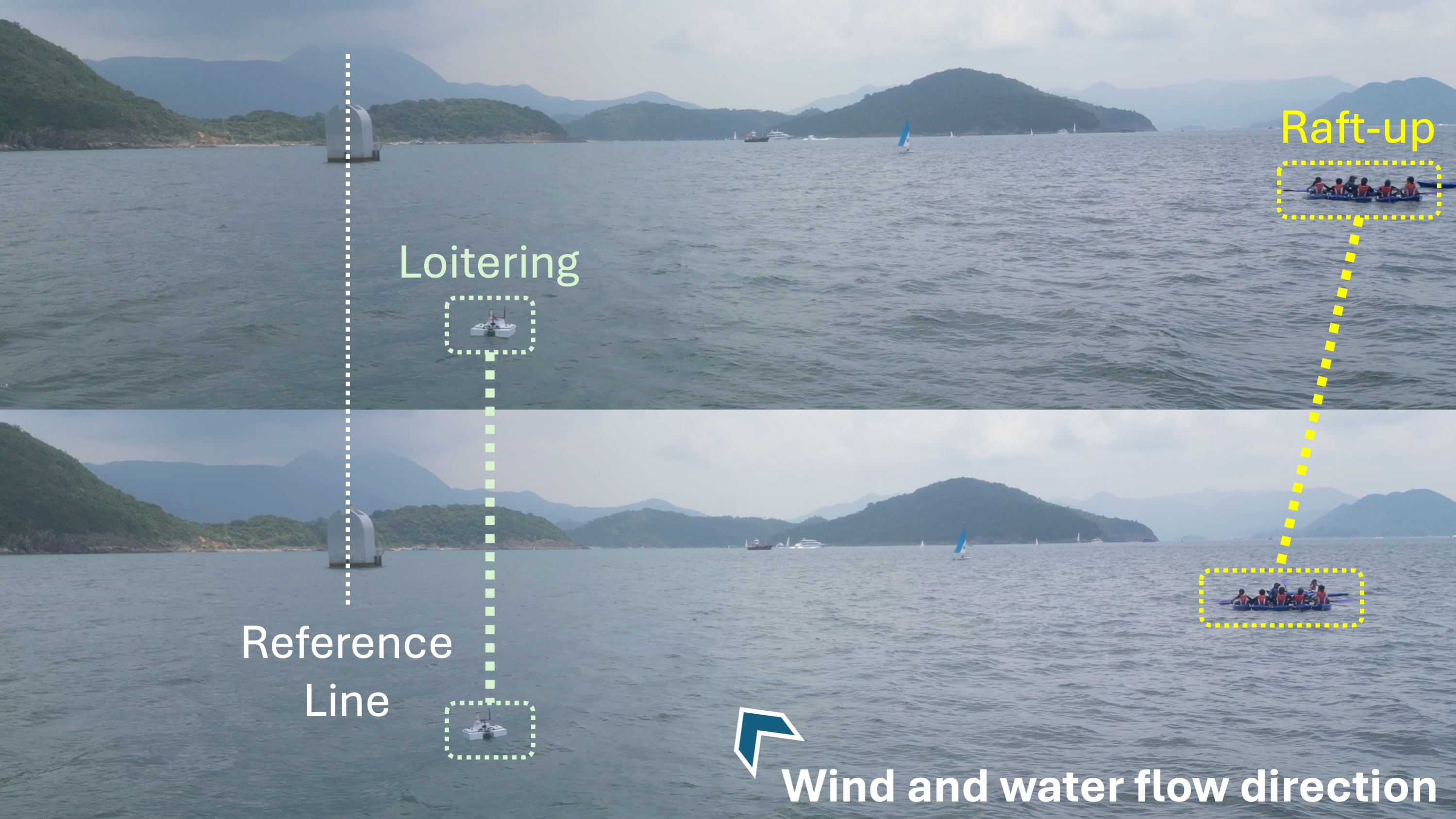}}
\caption{Experimental results of loiter mode for the ASV, showing its ability to maintain a fixed position in the presence of environmental disturbances.} 
\label{fig:loiter_mode} 
\end{figure}

The loiter mode test results, as shown in Fig. \ref{fig:loiter_mode}, illustrate the ASV’s performance in maintaining its position. The ASV was able to hold its position accurately, as indicated by the reference line and the minimal drift observed during the test, the environment condition is shown in Table \ref{table:wind speed and current on test day}. The loiter mode was particularly effective in keeping the ASV stationary, even in the presence of moderate waves and currents.
The loiter mode test confirmed the ASV’s capability to perform station-keeping tasks with high precision. This functionality is crucial for various applications, such as monitoring specific underwater areas, supporting other underwater vehicles, and conducting prolonged observations. The ability to maintain a stable position enhances the ASV’s operational flexibility and effectiveness in real-world scenarios.
\begin{table}[htbp]
\caption{Wind speed and current on test day.}
\begin{center}
\begin{tabular}{|>{\centering\arraybackslash}p{4cm}|>{\centering\arraybackslash}p{3cm}|}
\hline
\textbf{Beaufort Wind Scale} & 4          \\ \hline
\textbf{Mean wind speed}     & 20-30 km/h \\ \hline
\textbf{Height of waves}     & 0.5 m (Pier)       \\ \hline
\textbf{Surface Layer Flow}  & 0.15 km/h  \\ \hline
\end{tabular}
\label{table:wind speed and current on test day}
\end{center}
\end{table}

\subsection{BU-HEXA Operational Efficiency and Adaptability Test} The second experiment focused on evaluating the operational efficiency and adaptability of the BU-HEXA hexapod robots in various underwater terrains. The robots were tested in simulated environments, including sandy, rocky, and muddy substrates, to assess their locomotion capabilities and obstacle navigation. A comprehensive series of simulations were conducted using CoppeliaSim to determine the optimal walking stance, size, and component configuration for the hexapod robots. These simulations provided critical insights into minimizing drag resistance and enhancing operational efficiency.


Following the simulations, field tests were conducted in a controlled coastal environment. The hexapod robots were equipped with 3D-printed PETG panels to increase the contact area of their legs, preventing them from becoming embedded in sandy substrates. The robots’ performance was evaluated based on their ability to navigate over different terrains and maintain stability. The results demonstrated that the hexapod robots effectively navigated over rocky, sandy, and muddy terrains, maintaining stability and operational efficiency. The enhanced leg design significantly improved their mobility, preventing them from becoming stuck in sandy substrates.

\begin{figure}[htbp] 
\centerline{\includegraphics[width=1\linewidth]{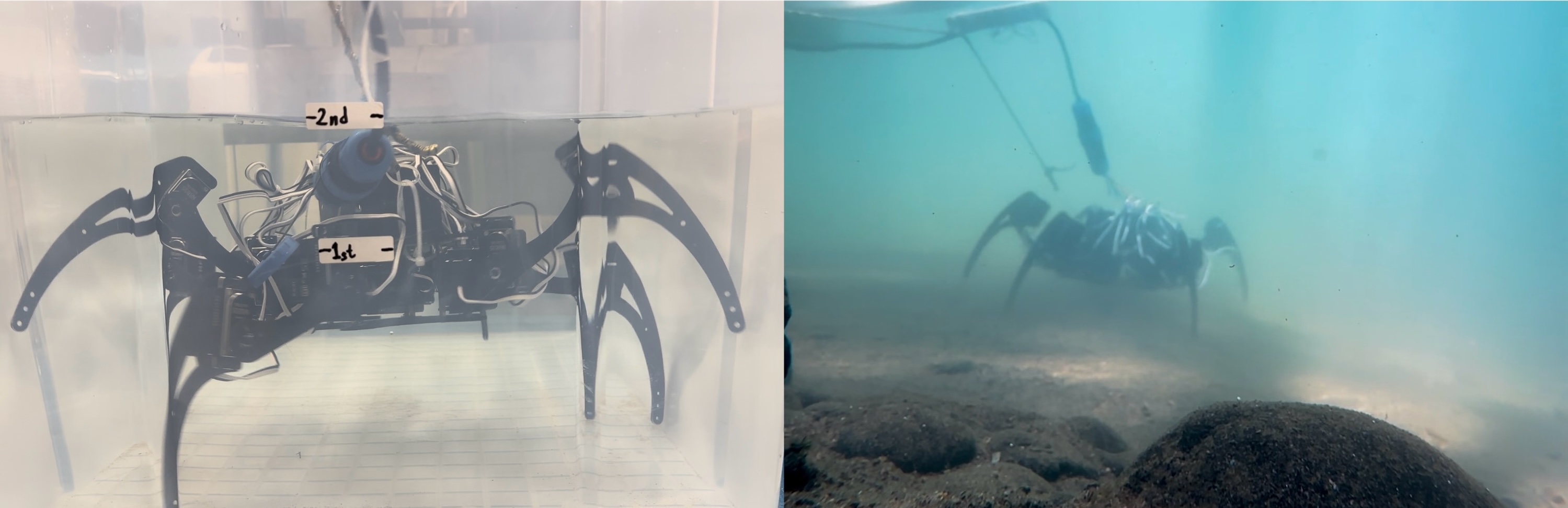}} 
\caption{BU-HEXA undergoing water tank test then deployed in a lake for real-world operational testing.}
\label{fig:BU-HEXA_trial} 
\end{figure}

To further evaluate the operational efficiency and waterproofing of the BU-HEXA, controlled tests were conducted in a water tank, as shown in Fig. \ref{fig:BU-HEXA_trial}. The hexapod robot was submerged and tested for its ability to maintain stability and function under controlled water conditions. This test also included the assessment of the robot’s movement and articulation in a confined aquatic environment. The BU-HEXA was placed in a water tank and its movements were monitored for any signs of water ingress or mechanical failure. The robot’s sensors and actuators were tested for responsiveness and accuracy while submerged. The test aimed to simulate real-world underwater conditions in a controlled setting to ensure the robustness and reliability of the BU-HEXA.
The results from the water tank test confirmed that the BU-HEXA could maintain stable operation and effective movement while submerged. There were no significant issues with water ingress, and all sensors and actuators functioned as expected. This test further validated the BU-HEXA’s design for underwater operations. To evaluate the BU-HEXA’s performance in a more realistic and dynamic environment, an additional field test was conducted in a lake, as shown in Fig. \ref{fig:BU-HEXA_trial}. The hexapod robot was deployed in the lake to assess its stability, mobility, and operational efficiency in natural underwater conditions. This test aimed to validate the robot’s ability to navigate and perform tasks in an uncontrolled environment, providing a closer approximation to potential real-world applications.

During the lake test, the BU-HEXA was monitored for its ability to traverse various underwater terrains, including areas with vegetation, mud, and rocks. The robot’s sensors and actuators were continuously monitored to ensure accurate data collection and responsive control. The real-time video feedback from the onboard camera was used to analyze the robot’s movements and interactions with the environment. The results from the lake test demonstrated the BU-HEXA’s robust performance in natural underwater conditions. The hexapod robot successfully navigated through different terrains and maintained stability and operational efficiency. The sensors and actuators performed reliably, providing accurate data and responsive control throughout the test. This field test further validated the BU-HEXA’s design and operational capabilities, confirming its potential for real-world underwater applications. Specific performance metrics were recorded to evaluate the effectiveness of the BU-HEXA in various conditions. Table \ref{table:performance_metrics} summarizes the key findings from the tests.
\begin{table}[htbp]
\caption{Performance Metrics of BU-HEXA.}
\begin{center}
\begin{tabular}{|>{\centering\arraybackslash}p{2.7cm}|>{\centering\arraybackslash}p{1.8cm}|>{\centering\arraybackslash}p{2.7cm}|}
\hline
\textbf{Metric} & \textbf{Value/Range} & \textbf{Test Conditions/Notes} \\
\hline
GPS Accuracy & < 2.5 meters & Multiple trials \\
\hline
ASV Speed & 2 m/s & Multiple trials \\
\hline
BU-HEXA Speed on Sand & 0.1-0.3 m/s & Controlled environment \\
\hline
BU-HEXA Speed on Rocky Terrain & 0.05-0.2 m/s & Controlled environment \\
\hline
\end{tabular}
\label{table:performance_metrics}
\end{center}
\end{table}
To calculate the searching area per hour based on the BU-HEXA’s speed, we consider the range of speeds the robot can achieve on different terrains: 0.1 to 0.3 m/s on sand and 0.05 to 0.2 m/s on rocky surfaces. We first convert the speed from meters per second (m/s) to meters per hour (m/h): Speed (m/h) = Speed (m/s) × 3600. For sand, this results in a minimum speed of 0.1 × 3600 = 360 m/h and a maximum speed of 0.3 × 3600 = 1080 m/h. For rocky surfaces, the minimum speed is 0.05 × 3600 = 180 m/h and the maximum speed is 0.2 × 3600 = 720 m/h. Assuming a 1-meter-wide search path, the area covered per hour on sand ranges from 360 m²/h to 1080 m²/h, and on rocky surfaces from 180 m²/h to 720 m²/h.

\subsection{TUV Imaging and Object Detection Test}
The TUV was tested for its imaging capabilities and real-time object detection. The underwater camera provided high-quality video feeds, which were processed using machine learning algorithms for object identification. The system successfully detected and categorized underwater objects, demonstrating its potential for evidence recovery operations. The TUV's performance in capturing high-resolution imagery and transmitting real-time video data was evaluated in various underwater conditions. The results showed that the TUV could effectively identify and classify objects, providing valuable data for evidence recovery and environmental monitoring. The integration of machine learning algorithms enhanced the system's ability to process and analyze video feeds, improving the accuracy and efficiency of the object detection process.

The TUV was also equipped with a robust tethering system to ensure reliable power and data transmission. This setup allowed continuous operation and real-time monitoring, essential for extended missions and complex underwater searches. The camera system was designed to withstand high-pressure environments, ensuring consistent performance in deeper waters.
\begin{figure}[htbp]
\centerline{\includegraphics[width=1\linewidth]{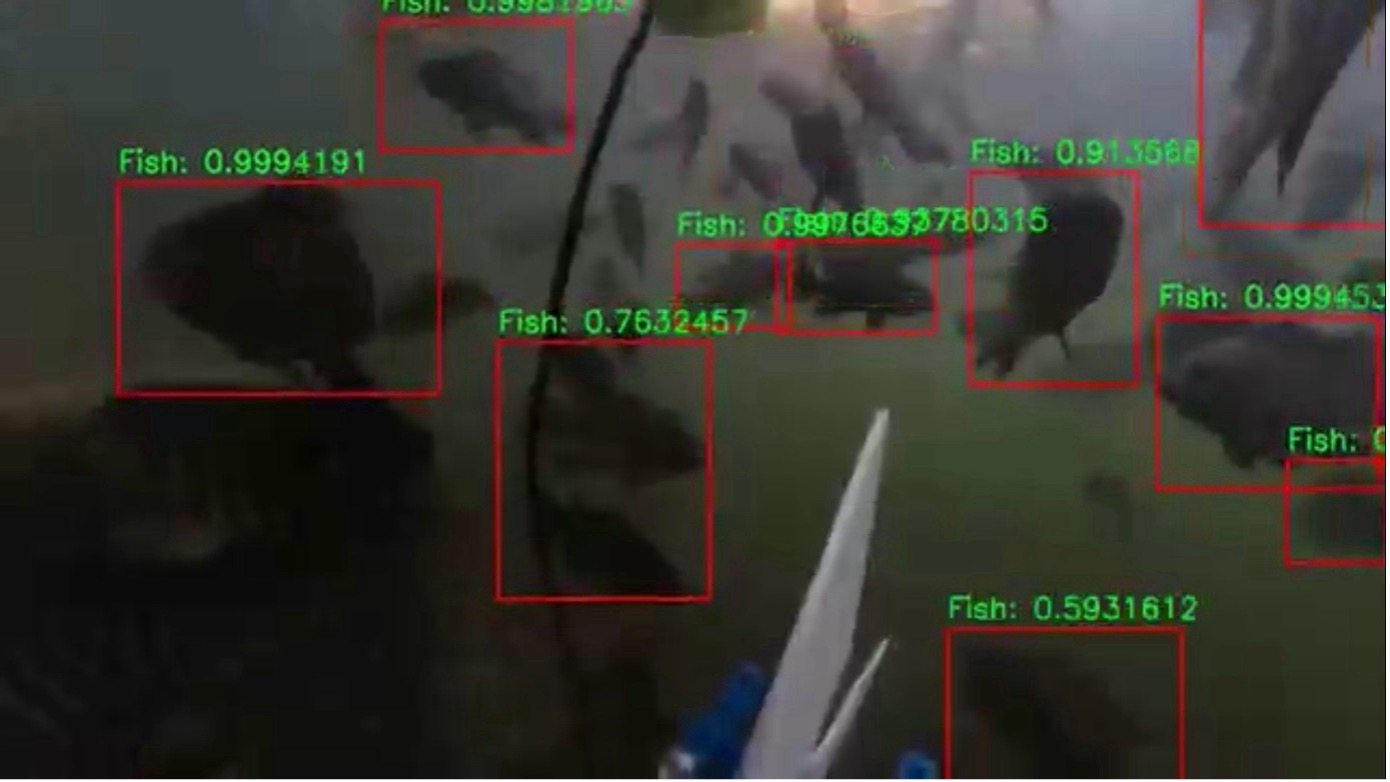}}
\caption{Object detection function work in underwater environment}
\label{fig:object_detection}
\end{figure}
The object detection capabilities of the TUV were further validated through tests in different water bodies, including ponds and coastal areas. The system demonstrated high accuracy in identifying various objects, such as simulated evidence items and natural underwater features. The machine learning model was trained using a diverse dataset, which included images from controlled environments and real-world scenarios. This approach ensured that the model could generalize well and perform reliably across different conditions.

The integration of advanced sensor technologies and machine learning algorithms significantly enhanced the TUV's capabilities. The system was able to provide real-time feedback, enabling operators to make informed decisions during search and recovery operations. The use of machine learning also reduced the time required for manual inspection, improving the overall efficiency of the process.

\section{Conclusion}
The CUES-S offers a robust and efficient solution for underwater evidence search and recovery. The combination of an autonomous surface vehicle, a biomimetic underwater hexapod, and a towed underwater vehicle provides a comprehensive approach to addressing the challenges of underwater operations. The system's design and experimental validation demonstrate its potential for real-world applications, including law enforcement and marine research.

Future work will focus on enhancing the system's autonomy, improving communication technologies, and conducting comprehensive environmental impact studies. The integration of advanced sensor technologies and machine learning algorithms will further enhance the system's capabilities, providing a valuable tool for law enforcement and marine research. By continuing to refine and expand the capabilities of the CUES-S, we aim to develop a versatile and reliable system that can address a wide range of underwater search and recovery challenges.

{\small
\bibliographystyle{ieeetr}
\bibliography{main}

\begin{thebibliography}{10}

\bibitem{qiu2019underwater}
T.~Qiu, Z.~Zhao, T.~Zhang, C.~Chen, and C.~P. Chen, ``Underwater internet of things in smart ocean: System architecture and open issues,'' {\em IEEE Transactions on Industrial Informatics}, vol.~16, no.~7, pp.~4297--4307, 2019.

\bibitem{plueddemann2011autonomous}
A.~Plueddemann, A.~Kukulya, R.~Stokey, and L.~Freitag, ``Autonomous underwater vehicle operations beneath coastal sea ice,'' {\em IEEE/ASME Transactions on Mechatronics}, vol.~17, no.~1, pp.~54--64, 2011.

\bibitem{rutledge2018intelligent}
J.~Rutledge, W.~Yuan, J.~Wu, S.~Freed, A.~Lewis, Z.~Wood, T.~Gambin, and C.~Clark, ``Intelligent shipwreck search using autonomous underwater vehicles,'' in {\em IEEE International Conference on Robotics and Automation (ICRA)}, pp.~6175--6182, 2018.

\bibitem{park2015multi}
J.-Y. Park, H.~Shim, H.~Baek, S.~Yoo, B.-H. Jun, and P.-M. Lee, ``Multi-legged rov crabster and an acoustic camera for survey of underwater cultural heritages,'' in {\em MTS/IEEE Oceans}, pp.~1--5, 2015.

\bibitem{pengpai_auv}
{Qingdao Pengpai Marine Exploration Technology Co., Ltd.}, 2024.

\bibitem{qu2024recent}
J.~Qu, Y.~Xu, Z.~Li, Z.~Yu, B.~Mao, Y.~Wang, Z.~Wang, Q.~Fan, X.~Qian, M.~Zhang, {\em et~al.}, ``Recent advances on underwater soft robots,'' {\em Advanced Intelligent Systems}, vol.~6, no.~2, p.~2300299, 2024.

\bibitem{fallon2010cooperative}
M.~F. Fallon, G.~Papadopoulos, J.~J. Leonard, and N.~M. Patrikalakis, ``Cooperative auv navigation using a single maneuvering surface craft,'' {\em The International Journal of Robotics Research}, vol.~29, no.~12, pp.~1461--1474, 2010.

\bibitem{wang2023cooperative}
Y.~Wang, W.~Liu, J.~Liu, and C.~Sun, ``Cooperative usv--uav marine search and rescue with visual navigation and reinforcement learning-based control,'' {\em ISA Transactions}, vol.~137, pp.~222--235, 2023.

\bibitem{wang2022quadrotor}
P.~Wang, C.~Wang, J.~Wang, and M.~Q.-H. Meng, ``Quadrotor autonomous landing on moving platform,'' {\em Procedia Computer Science}, vol.~209, pp.~40--49, 2022.

\bibitem{lindsay2022collaboration}
J.~Lindsay, J.~Ross, M.~L. Seto, E.~Gregson, A.~Moore, J.~Patel, and R.~Bauer, ``Collaboration of heterogeneous marine robots toward multidomain sensing and situational awareness on partially submerged targets,'' {\em IEEE Journal of Oceanic Engineering}, vol.~47, no.~4, pp.~880--894, 2022.

\bibitem{li2021development}
J.~Li, T.~Chen, Z.~Yang, L.~Chen, P.~Liu, Y.~Zhang, G.~Yu, J.~Chen, H.~Li, and X.~Sun, ``Development of a buoy-borne underwater imaging system for in situ mesoplankton monitoring of coastal waters,'' {\em IEEE Journal of Oceanic Engineering}, vol.~47, no.~1, pp.~88--110, 2021.

\bibitem{koutalakis2022river}
P.~Koutalakis and G.~N. Zaimes, ``River flow measurements utilizing uav-based surface velocimetry and bathymetry coupled with sonar,'' {\em Hydrology}, vol.~9, no.~8, p.~148, 2022.

\end{thebibliography}
}

\end{document}